\documentclass[
twocolumn,
]{ceurart}

\usepackage{latexsym}
\usepackage[utf8]{inputenc}
\usepackage{microtype}
\usepackage{inconsolata}
\usepackage{graphicx}
\usepackage{changepage}
\usepackage{hyperref}
\usepackage{caption}
\usepackage{subcaption}
\usepackage{multirow}
\usepackage{tikz}
\usepackage{enumitem}
\usepackage[utf8]{inputenc}

\newcommand\T{\rule{0pt}{3ex}}          
\newcommand\Tsxx{\rule{0pt}{2ex}}       

\usepackage{listings}
\begin{document}

\copyrightyear{2023}
\copyrightclause{Copyright for this paper by its authors.
  Use permitted under Creative Commons License Attribution 4.0
  International (CC BY 4.0).}

\conference{RecSys in HR'23: The 3rd Workshop on Recommender Systems for Human Resources, in conjunction with the 17th ACM Conference on Recommender Systems, September 18--22, 2023, Singapore, Singapore.}

\title{Résumé Parsing as Hierarchical Sequence Labeling: An Empirical Study}

\author[1]{Federico Retyk}[
    email=machinelearning@avature.net
]
\author[1]{Hermenegildo Fabregat}[]
\author[1]{Juan Aizpuru}[]
\author[1]{Mariana Taglio}[]
\author[1]{Rabih Zbib}[]
\address[1]{Avature Machine Learning}

\begin{abstract}
Extracting information from résumés is typically formulated as a two-stage problem, where the document is first segmented into sections and then each section is processed individually to extract the target entities. Instead, we cast the whole problem as sequence labeling in two levels ---lines and tokens--- and study model architectures for solving both tasks simultaneously. We build high-quality résumé parsing corpora in English, French, Chinese, Spanish, German, Portuguese, and Swedish. Based on these corpora, we present experimental results that demonstrate the effectiveness of the proposed models for the information extraction task, outperforming approaches introduced in previous work. We conduct an ablation study of the proposed architectures. We also analyze both model performance and resource efficiency, and describe the trade-offs for model deployment in the context of a production environment.
\end{abstract}


\maketitle


\section{Introduction}
\label{sec:introduction}

Résumé parsing has become increasingly important in the context of digitalized recruitment processes. It involves the extraction of relevant information about a candidate from their résumé document into a structured data model. The extracted information, when integrated with downstream recommender systems, can in turn help candidates and recruiters optimize their search.

Résumés\footnote{We use the term résumé synonymously with curriculum vitæ (CV).} exhibit significant diversity in their format and in their use of language due to differences in the background, industry, and location of the candidates~\cite{tosik-etal-2015-word}. But despite their unstructured nature, these documents are typically organized into sections. Each one of these text blocks contains important details about the candidate, such as their personal information, education history, previous work experience, and professional skills. Additionally, some sections depict a chronological progression (e.g.~work experience, education), and are naturally further divided into groups. Within particular sections and groups are different concepts or entities that are relevant to the recruitment process. For example, the full name of the candidate, their address, and their phone number are typically present in the \textit{contact information} section. Each group within the \textit{work experience} section usually includes a period, a job title, and the employer's name.

Since résumé parsing occurs early on in the digitalized recruitment process pipeline, its accuracy has a significant effect on that of the downstream recommender systems. However, the aforementioned diversity in the use of language in résumés makes the parsing problem challenging for pattern matching or other classical artificial intelligence (AI) approaches. Effective solutions for this task, therefore, require the use of machine learning techniques.

Existing industry-scale solutions for résumé parsing do not make public detailed information about their systems. On the other hand, previous academic research in this domain focuses on constrained scenarios that are limited in scope, in the complexity of the target label scheme, or in terms of the size and quality of the annotated datasets. Moreover, these works address the problem in two or more stages. In the first stage, they segment the résumé into sections and groups~\cite{tosik-etal-2015-word, zu2019textblock, barducci2022ir}. Since résumés are long text documents, this is generally approached as text classification of independent lines without document-level context. The second stage uses a section-specific sequence labeling model to extract the target entities from the text of each section.

In this work, we propose a joint model that labels the full document as a whole. This is an unusual setting in academic literature for sequence labeling, as résumés are long text sequences and the set of labels is relatively big. We show that the proposed system is not only efficient and convenient from an engineering point of view, but it is also competitive with the two-stage alternative. We compare it to previous approaches and we also study several design-decisions of our system in terms of their effect on accuracy as well as in time and memory efficiency. We share experimental observations on résumés in seven languages and provide insight into the deployment of this system in production environments. In summary, the main contributions of this paper are:

\begin{itemize}
    \item Casting the task of résumé parsing as hierarchical sequence labeling, with line-level and token-level objectives, and presenting an efficient résumé parsing architecture for simultaneous labeling at both levels. We propose two variants of this model: one optimized for latency and the other optimized for performance.
    \item A comprehensive set of experiments on résumé parsing corpora in English, French, Chinese, Spanish, German, Portuguese, and Swedish, each covering diverse industries and locations. We share our experience in the process of developing such annotations. These experiments compare our proposed system to previous approaches and include an extensive ablation study, examining various design choices of the architecture.
    \item Insights into the process of deploying this model in a global-scale production environment, where candidates and recruiters from more than 150 countries use it to parse over 2 million résumés per month in all these languages. We analyze the trade-off between latency and performance for the two variants of the model we propose.
\end{itemize}

Our empirical study suggests that the proposed hierarchical sequence labeling model can parse résumés effectively and outperform previous work, even with a task definition that involves labeling significantly large text sequences and a relatively large number of entity labels.


\section{Related Work}
\label{sec:related-work}

Our work builds upon prior research on deep learning for sequence labeling, specifically those applying neural networks in combination with Conditional Random Fields (CRFs) to various sequence labeling tasks. Huang et al.~(2015) investigated an architecture based on Bidirectional Recurrent Neural Networks (BiRNNs) and CRFs~\cite{huang2015bidirectional}. They use both word embeddings and handcrafted features as initial representations. Lample et al.~(2016) extended this architecture by introducing character-based representations of tokens as a third source of information for the initial features~\cite{lample-etal-2016-neural}. An alternative character-based approach was proposed by Akbik et al.~(2018), which uses a BiRNN over the character sequence to extract contextualized representations that are then fed to a token-level BiRNN+CRF~\cite{akbik-etal-2018-contextual}. In addition, Devlin et al.~(2019) introduce a simple Transformer-based approach that avoids the utilization of CRF. This consists of a pre-trained BERT encoder, which is fine-tuned, followed by a linear classification layer applied to the representation of each token~\cite{devlin-etal-2019-bert}. We refer interested readers to the surveys by Yadav and Bethard~(2018) and Li et al.~(2022) for a more comprehensive review of deep neural networks for sequence labeling~\cite{yadav-bethard-2018-survey, li2022survey}.

Prior work on parsing résumés usually divides the problem into two tasks, and tackles each separately~\cite{tosik-etal-2015-word, zu2019textblock, barducci2022ir, ayishathahira2018crf, sajid2022resume}. The résumé is first segmented into sections and groups, and then section-specific sequence labeling models are applied to extract target entities. The early work by Tosik et al.~(2015) focuses on the second task only, as they experiment with already-segmented German résumés~\cite{tosik-etal-2015-word}. They train named entity recognition models for the \textit{contact information} and \textit{work experience} sections, each with a small set of labels. The architecture they apply uses word embeddings as direct features for the CRF.

Zu et al.~(2019) use a large set of English résumés collected from a single Chinese job board to experiment with several architectures for each of the two stages~\cite{zu2019textblock}. For segmentation, they classify each line independently (without document context). Then to extract entities, they train different models for each section type. The input to these sequence labeling models is the text of each independent line. While for the line classification task they use manually annotated samples, the sequence labeling models are trained using automatic annotations based on gazetteers and dictionaries.

Barducci et al.~(2022) work with Italian résumés. They first segment the résumé using a pattern-matching approach that relies on a language- and country-specific dictionary of keywords~\cite{barducci2022ir}. After this, they train independent sequence labeling models for each section type. The architecture they use for the sequence labeling component is based on the approach described above that uses BERT~\cite{devlin-etal-2019-bert} with a classification layer on top.

Finally, Pinzon et al.~(2020) work with a small corpus of English résumés~\cite{pinzon2020ner}. They bypass the segmentation task (ignoring sections and groups) and propose a model that directly extracts entities from the résumé text. They use a BiRNN+CRF model for the token-level sequence labeling task. Among the related work we examined, this is the only one that made their dataset public. Nevertheless, a manual examination of the corpus led us to conclude that the sample is far from representative of real-world English résumés and that the labeling scheme they use is limited and inadequate for our scope.

We extend the previous work by exploring a joined architecture that predicts labels for both lines and tokens, treating each as a sequence labeling task. Furthermore, as in Pinzon et al.~(2020)~\cite{pinzon2020ner}, we unify the extraction of entities for any section. This setup is challenging, since résumés are unusually long compared to typical Information Extraction tasks, and the set of labels for entities is also bigger. But the advantage is the improvement of efficiency in terms of execution time and memory usage, and the simplification of the engineering effort since only one model needs to be trained, deployed, and maintained. 

Our work is also the first one to study résumé parsing in seven languages, with large corpora of résumés selected from many different industry sectors, and using high-quality manual annotations for both the line and token tasks.


\section{Task Description}
\label{sec:task}

We cast résumé parsing as a hierarchical sequence labeling problem, with two levels: the line-level and the token-level. These two tasks can be tackled either sequentially or in parallel.

For the first, we view the résumé as a sequence of lines and infer the per-line labels that belong to different section and group types. This is a generalization of the task definition used in previous work, where the label (class) for each line is inferred independent of information about the text or the predicted labels of other lines. We assume that section and group boundaries are always placed at the end of a line, which is the case in all the résumés we came across during this project. The label set for this part of the task includes a total of 18 sections and groups, which are listed in Appendix~\ref{subsec:appendix-guidelines-lines}.

For the second level, we view the résumé as a long sequence of tokens that includes all the tokens from every line concatenated together. We infer the per-token labels that correspond to the different entities. The label set for this part of the task includes 17 entities, which are in turn listed in Appendix~\ref{subsec:appendix-guidelines-tokens}.

The scope of this paper revolves around the extraction task and therefore we do not focus on the conversion of the original résumé (e.g.~a \texttt{docx} or \texttt{pdf} file) into plain text format. Rather, the systems studied in this work assume textual input.


\section{Corpora}
\label{sec:corpora}

We built résumé parsing corpora in English, French, Chinese, Spanish, German, Portuguese, and Swedish. Some statistics on the corpora are reported in Table~\ref{tab:corpora-main}. For each of these languages, résumés were randomly sampled from public job boards, covering diverse locations and industries. For all but Chinese, we controlled the sampling process in order to enforce diversity in locations. For example, although the English corpus is biased toward the USA, there is a fraction of résumés from other English-speaking countries including the UK, Ireland, Australia, New Zealand, South Africa, and India. Although we did not control for industry variability, we observe a high level of diversity in the selected collections. We then used third-party software to convert into plain text the original files, which came in varied formats such as \texttt{pdf}, \texttt{doc}, and \texttt{docx}.

Since this effort is aimed at building a real-world application, annotation quality is highly important. For that purpose, we implemented a custom web-based annotation tool that allows the user to annotate section and group labels for each line of a résumé, and to annotate entity labels for each arbitrary span of characters. 

We developed the annotation guidelines by starting with a rough definition for each label and performing exploratory annotations on a small set of English résumés ---a mini-corpus that we later used for onboarding the annotators. The guidelines were then iteratively refined for the whole duration of the project, achieving a stable and rigorous version at the end. In Appendix~\ref{sec:appendix-guidelines} we define the section, group, and entity objectives covered in our corpora, and we provide a screenshot of the annotation tool user interface for reference.

Each language corpus was managed as an independent annotation project. We recruited 2 or 3 annotators, who are native speakers of the target language and without specifically seeking domain expertise, through an online freelance marketplace. The annotators did not communicate with each other during the process, maintaining the independence of the multiple annotations. Before starting the annotations on the target corpus, we asked each annotator to carefully read the guidelines, and annotate the onboarding English mini-corpus. After reviewing and providing feedback, the annotator was instructed to annotate all the résumés in the target corpus.

The estimated inter-annotator agreement~(IAA) for the corpus in each language, computed as suggested by Brandsen et al.~(2020)~\cite{brandsen-etal-2020-creating} in terms of $F_{1}$, ranges from 84.23 to 94.35\% and the median is 89.07\%.
Finally, we adjudicated the independent annotations in order to obtain the gold standard annotations. This process involved resolving any conflicting decisions made by individual annotators through the majority voting method. In cases where a majority decision was not attainable, the adjudicator was instructed to review the decisions of each annotator and apply their own criteria to arrive at a final decision.


\begin{table}
\begin{center}
{\caption{The number of résumés and the average number of lines and tokens per résumé for each language corpus.}\label{tab:corpora-main}}
\begin{tabular}{lccc}
\hline
\textbf{Corpus} & \textbf{Résumés} & \textbf{Lines} & \textbf{Tokens} \\
\hline
English & 1196 & 73.3 & 834.1\\
French & 1044 & 54.4 & 539.1\\
Chinese & 1023 & 50.6 & 664.8\\
Spanish & 846 & 68.6 & 667.4\\
German & 738 & 80.5 & 608.6\\
Portuguese & 628 & 73.1 & 773.6\\
Swedish & 519 & 74.5 & 632.0\\
\hline
\end{tabular}
\end{center}
\end{table}


\section{Model Architecture and Training}
\label{sec:model}

\begin{figure*}[t]
\subfloat[Single-task variant for sequence labeling on lines.]{\label{fig:architecture-a}\includegraphics[width=0.92\textwidth]{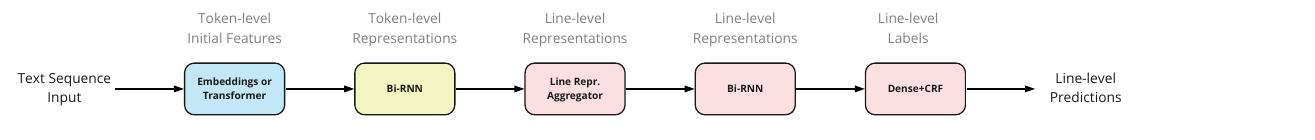}} \\ \par\bigskip
\subfloat[Single-task variant for sequence labeling on tokens.]{\label{fig:architecture-b}\includegraphics[width=0.92\textwidth]{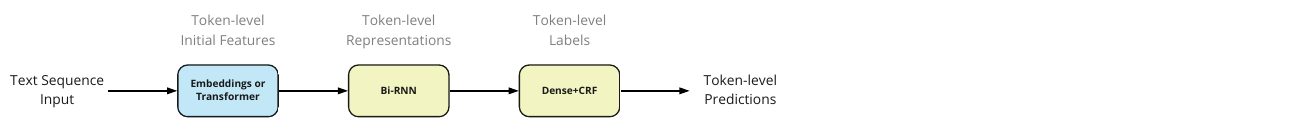}} \\ \par\bigskip
\subfloat[Multi-task variant for sequence labeling on both lines and tokens.]{\label{fig:architecture-c}\includegraphics[width=0.92\textwidth]{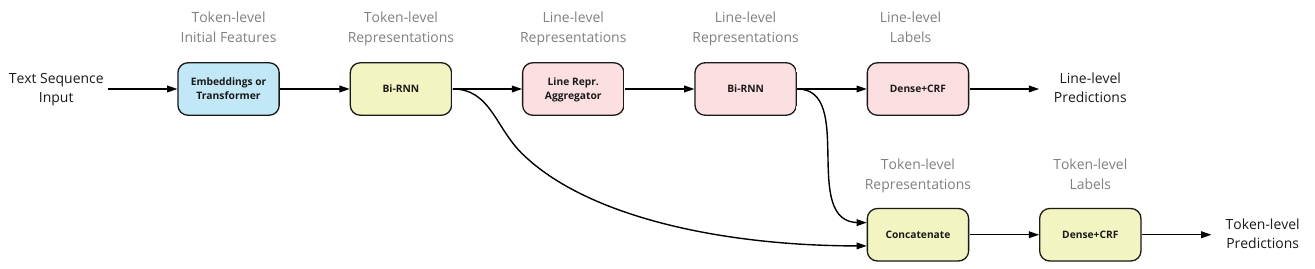}}
\caption{Model variants studied in this work. Blue architecture blocks denote initial features (which are pre-trained, and held fixed during our experiments), while yellow and red blocks denote layers that output a sequence of elements for each token or line.}
\centering
\label{fig:architecture}
\end{figure*}

The models we use in this work are based on the BiRNN+CRF architecture. Initial features are first extracted for each token, then combined through bidirectional recurrent layers, and finally passed through a CRF layer to predict the labels. Unless specified otherwise, the input to the model is the entire résumé text after applying tokenization. We study two design-decisions: (1) the choice for initial features, and (2) separate models for predicting line and token labels vs.\ a multi-task model that predicts both jointly.

\noindent \textbf{Initial features}. We explore two alternatives: 

\begin{enumerate}[label=(\alph*)]
\item A combination of FastText~\cite{bojanowski-etal-2017-enriching} word embeddings and handcrafted features, which are detailed in Appendix~\ref{sec:handcrafted}.
\item Token representations obtained from the encoder component of a pre-trained T5~\cite{raffel2020exploring} model (or an mT5~\cite{xue-etal-2021-mt5}, depending on the language) without fine-tuning. 
\end{enumerate}

\begin{figure}[t]
\centering
\includegraphics[width=0.47\textwidth]{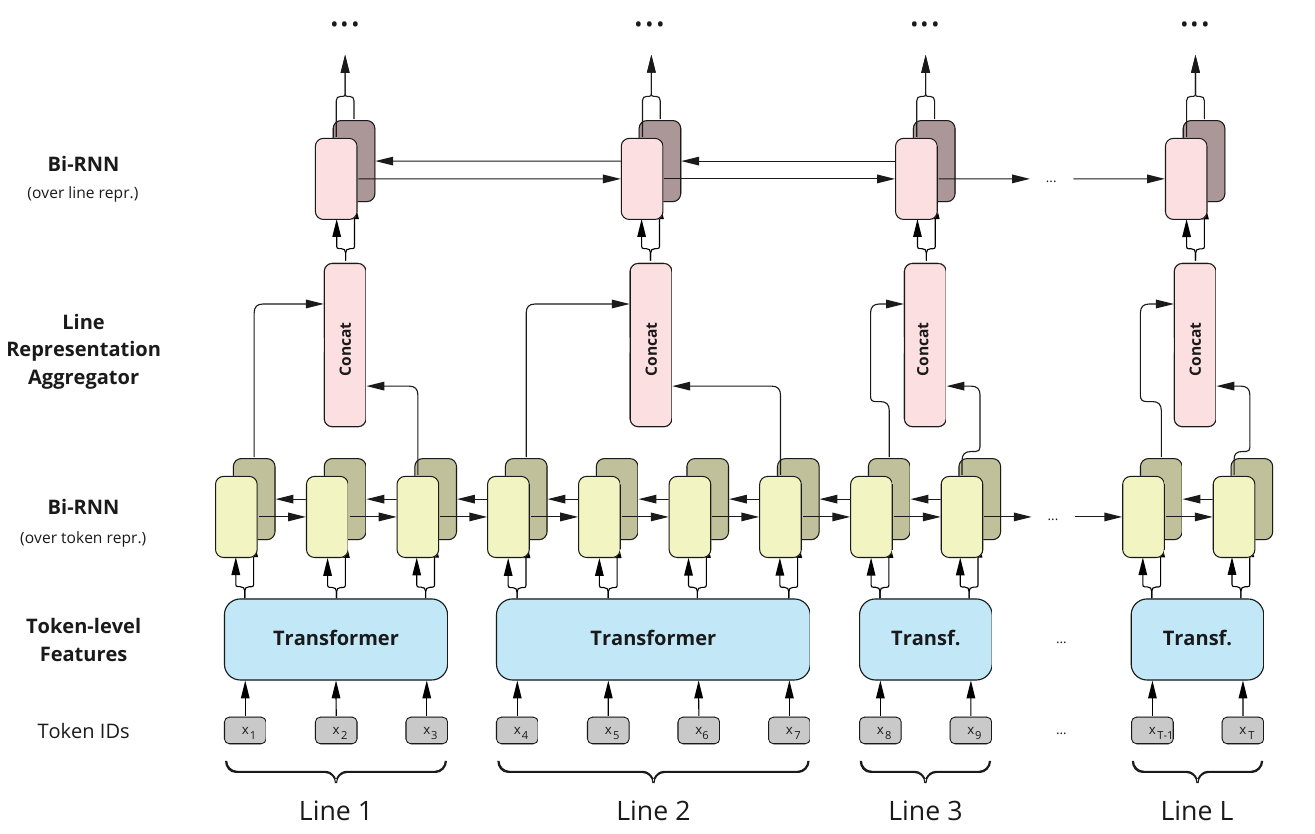}
\caption{Detail of the aggregation of token-level representations into line-level representations (blocks in red), exemplified with a variant using Transformer-based initial features. The BiRNN that contextualizes initial token-level features across every line (blocks in yellow) is needed because a typical résumé does not fit in the maximum input length of the typical Transformer models. }
\label{fig:architecture-detail}
\end{figure}

The T5 models are based on the Transformer~\cite{vaswani2017attention} architecture. For this second case, each line is encoded individually\footnote{Note that résumés are long text sequences, usually longer than 512 tokens (see Table~\ref{tab:corpora-main}).}, and then the token representations for each line are concatenated to obtain the input sequence for the BiRNN+CRF architecture. This is visually described in Figure~\ref{fig:architecture-detail}. Preliminary experiments, which are not presented here because of space constraints, showed that avoiding the BiRNN component for this last architecture, i.e.\ applying CRF directly on the output of the Transformer-based features, obtains markedly worse results. This is because the two layers capture complementary aspects of the context: the Transformer encodes tokens by exclusively considering the context of the current line, while the BiRNN layer on top contextualizes across every line. Because of the typical length of a résumé in terms of tokens, we did not explore encoding the whole résumé at once with the Transformer encoders used in this work.

\noindent \textbf{Single-task vs.\ Multi-task}. We experiment with: 

\begin{enumerate}[label=(\alph*)]
\item Single-task models that perform either line-level sequence labeling (sections and groups) or token-level sequence labeling (entities).
\item Multi-task models that predict labels for both line-level and token-level tasks simultaneously. 
\end{enumerate}

Figure~\ref{fig:architecture} illustrates the model variants. The architecture shown in Figure~\ref{fig:architecture-a} is a single-task model for line-level objectives (sections and groups). This architecture takes as input the complete sequence of tokens in the résumé and predicts one label for each line. We train this type of model using only the supervision from the line sequence labeling task. As the diagram shows, a sequence of token representations is transformed into a sequence of line representations, such that the output is expressed in terms of lines, using a pooling operation inspired by Akbik et al.~(2018)~\cite{akbik-etal-2018-contextual}. In detail, consider the input résumé as a long sequence $\mathbf{X} = (\mathbf{x}_1, \mathbf{x}_2, \dots, \mathbf{x}_T)$ of $T$ tokens, partitioned in $L$ lines. Each line $j$ is a subsequence of tokens, starting at $\mathbf{x}_{a_j}$ and ending at $\mathbf{x}_{b_j}$. After extracting the initial features for each token, and feeding these into the token-wise BiRNN layer, we obtain a sequence of token representations $\mathbf{H} = (\mathbf{h}_1, \mathbf{h}_2, \dots, \mathbf{h}_T)$, each consisting of a forward and backward component, $\mathbf{h}_i = \overrightarrow{\mathbf{h}}_i \oplus \overleftarrow{\mathbf{h}}_i$. We then compute the representation for each line $j$ by concatenating the forward component of the last token with the backward component of the first token: $\mathbf{r}_j = \overrightarrow{\mathbf{h}}_{b_{j}} \oplus \overleftarrow{\mathbf{h}}_{a_{j}}$. The result is a sequence of line representations $\mathbf{R} = (\mathbf{r}_1, \mathbf{r}_2, \dots, \mathbf{r}_L)$, which is in turn processed by another BiRNN layer. This aggregation mechanism is depicted in Figure~\ref{fig:architecture-detail}.

Figure~\ref{fig:architecture-b}, on the other hand, shows the single-task model for token-level objectives (entities). This second architecture is trained using supervision from the token-level labels only.

Finally, a multi-task architecture for predicting both line and token objectives jointly is presented in Figure~\ref{fig:architecture-c}. It is trained with both supervision signals simultaneously. For this multi-task architecture, the token-level CRF receives as input the concatenation of: (i)~the representation of the target token and (ii)~the line-level representation of the line in which the token occurs.
All the models are implemented using TensorFlow~\cite{abadi2016tf}.


\begin{table*}
    {\caption{Performance of the model variants for résumé parsing in seven languages, expressed as micro-average $F_{1}$ score in percentage points for the positive labels in the two hierarchical levels of the sequence labeling task: token and lines objectives. For each variant, we report the average of three independent replications using different random seeds. (The single-task model for tokens using FastText features is equivalent to the one proposed by Pinzon et al.~(2020)~\cite{pinzon2020ner}.)}\label{tab:results-main}}
    \centering
    \begin{subtable}{\textwidth}
    \centering
    \begin{tabular}{lccccccc}
    \hline
    \Tsxx \textbf{Model} &
        \textbf{English} &
        \textbf{French} &
        \textbf{Chinese} &
        \textbf{Spanish} &
        \textbf{German} &
        \textbf{Portuguese} &
        \textbf{Swedish} \\
    \hline
    \Tsxx \textit{FastText features} & & & & & & & \\
    ~~Single-task (only tokens)&
        88.24 &  
        86.65 &  
        92.30 &  
        88.93 &  
        86.97 &  
        89.14 &  
        89.07 \\ 
    ~~Multi-task (lines and tokens) &
        89.03 &  
        86.90 &  
        92.66 &  
        88.77 &  
        87.43 &  
        89.14 &  
        89.14 \\ 
    \T \textit{Transformer features} & & & & & & & \\
    ~~Single-task (only tokens)&
        90.78 &  
        88.37 &  
        92.35 &  
        89.66 &  
        85.81 &  
        89.49 &  
        80.98 \\ 
    ~~Multi-task (lines and tokens) &
        90.94 &  
        88.65 &  
        92.61 &  
        90.24 &  
        86.13 &  
        90.05 &  
        81.15 \\ 
    \hline
    \end{tabular}
    \caption{Results for the token sequence labeling task (entities).}
    \label{tab:results-goals-main}
    \end{subtable}
    \par\bigskip
    \begin{subtable}{\textwidth}
    \centering
    \begin{tabular}{lccccccc}
    \hline
    \Tsxx \textbf{Model} &
        \textbf{English} &
        \textbf{French} &
        \textbf{Chinese} &
        \textbf{Spanish} &
        \textbf{German} &
        \textbf{Portuguese} &
        \textbf{Swedish} \\
    \hline
    \Tsxx \textit{FastText features} & & & & & & & \\
    ~~Single-task (only lines)&
       92.26  &  
       94.47  &  
       91.42  &  
       94.30  &  
       92.47  &  
       91.61  &  
       82.60  \\ 
    ~~Multi-task (lines and tokens) &
       92.24  &  
       94.30  &  
       91.54  &  
       93.75  &  
       91.55  &  
       90.58  &  
       83.14  \\ 
    \T \textit{Transformer features} & & & & & & & \\
    ~~Single-task (only lines)&
       90.95  &  
       84.99  &  
       85.76  &  
       81.32  &  
       87.09  &  
       89.02  &  
       58.37  \\ 
    ~~Multi-task (lines and tokens) &
       92.62  &  
       92.21  &  
       89.95  &  
       92.11  &  
       87.02  &  
       90.96  &  
       90.38  \\ 
    \hline
    \end{tabular}
    \caption{Results for the line sequence labeling task (sections and groups).}
    \label{tab:results-sections-main}
    \end{subtable}
\end{table*}


\section{Experiments}
\label{sec:experiments}

We next describe the results of our experiments using the corpora of Section~\ref{sec:corpora}. The main results are summarized in Table~\ref{tab:results-main}. For each language, we use 90\% of the documents for training and report the micro-average $F_{1}$ scores (for the positive labels only) on the held-out 10\%\footnote{Due to the relatively small size of the corpora, we opted against using a three-way split involving training, validation, and test sets.}. The results compare the two model architectures discussed in Section~\ref{sec:model}: {\em Single-task} and {\em Multi-task}, and for each architecture, the two alternatives for initial features: FastText and Transformer-based T5.
    
The $F_{1}$ scores for the token sequence labeling task (predicting entities) are reported in Table~\ref{tab:results-goals-main}. Those include the results for the two single-task models that act only on the token-level task, as well as the two multi-task models. The $F_{1}$ scores for the line sequence labeling task (sections and groups) are shown in Table~\ref{tab:results-sections-main}, again for the two single-task models that act only on the line-level task, and the two multi-task models\footnote{Row 2 of both sub-tables evaluates the same underlying model (but for different tasks), and similarly for row 4}.

We make some observations. Comparing row 1 with row 3, and also row 2 with row 4, we see that using Transformer-based embeddings yields an improvement of 2.5\% in the goals $F_{1}$ on English, and a smaller improvement on French, Spanish, Chinese, and Portuguese, but is worse on German and Swedish\footnote{Swedish is an outlier, where the Transformer-based models are markedly less accurate. This might be due to the small size of Swedish data used for pre-training mT5.}. FastText initial features, on the other hand, perform as well or better than Transformer-based features in the line-level task. It is important to consider, though, that the improved error rate of the Transformer-based model comes at a higher computational cost during inference. This consideration is especially important when the model is deployed in a high-load commercial application where latency is a crucial factor.

A second important observation is that the multi-task models generally outperform their single-task counterparts for the token sequence labeling task. Additionally, the multi-task model has a significant advantage in a commercial setting. From an operational perspective, the training, testing, integration, and maintenance of a single model is simpler and cheaper than for two models.


\subsubsection*{Section-specific Models}

The simplification of model development and maintenance is even more significant when we contrast the unified multi-task model described above with the typical two-stage approach for résumé parsing~\cite{tosik-etal-2015-word, zu2019textblock, barducci2022ir}. The latter requires training and maintaining several models: one for the initial line segmentation task, and then one for entity extraction within each specific section type (e.g.~one single-task model for the entities related to contact information, another single-task model for entities related to work experience, etc). By contrast, the unified multi-task model we proposed is used to label all the entities across the whole résumé at once, regardless of the section type. This simplification, however, comes at a cost of increased error rate since a section-specific model has to decide among a much smaller set of labels, and receives a shorter text sequence as input.

In this part, we attempt to quantify such degradation. We train section-specific models, i.e.~individual models, for the entities for three of the section types: \textit{contact information}, \textit{work experience}, and \textit{education}. Each is trained and evaluated only on the corresponding segment of the résumés. Segmentation is performed using the gold standard annotations for sections, in order to focus our measurements on the token-level task. In Table~\ref{tab:results-goals-separate}, we report the micro-average $F_{1}$ scores grouped by the relevant sections, comparing the performance of each section-specific model to the proposed unified, multi-task model. Results are reported for English, French, and Chinese.

We show a loss in $F_{1}$ ranging from 1\% to 5\% depending on the section and language. Since the section-specific models benefit from the gold standard segmentation of sections, the results should be considered as an upper bound of the degradation in error rate. A real-world system implemented according to the two-stage approach should expect a compound error carried from the first stage, e.g.~the error observed for the Single-task models presented in Table~\ref{tab:results-sections-main}. The aim is to provide the practitioner with a quantifiable assessment of the trade-off between engineering simplicity and task accuracy.


\subsubsection*{Analysis and Details on Deployment}

The results already suggest that the Transformer-based initial features perform generally better for the token-level sequence labeling task. Furthermore, they do not need language-specific handcrafted features, so they can be readily applied to new languages. On the other hand, the alternative set of initial features (the combination of word embeddings and handcrafted features) performs better in the line sequence labeling task for detecting section and group labels.

However, in terms of efficiency, our experiments reveal that using word embedding initial features leads to a considerable improvement in time-efficiency during inference, when compared to the Transformer-based features. The inference time for the multi-task model was measured under both feature sets. On a bare-metal server with a single GPU\footnote{NVIDIA Tesla T4 and Intel® Xeon® Platinum 8259CL CPU @ 2.50GHz.}, we observed a speedup of 7 of the FastText models compared to Transformer-based features. Furthermore, when utilizing CPU-only hardware\footnote{Intel® Xeon® CPU E5-2630 v2 @ 2.60GHz}, the speedup increased substantially to 90. As an example, we note that the multi-task model using FastText initial features, deployed on CPU-only servers via TensorFlow Serving~\cite{olston2017tfserving}, yields a latency of 450 ms per résumé without batch processing.


\begin{table*}
    \setlength{\tabcolsep}{5pt}
    {\caption{Comparison between the multi-task joint model and the single-task section-specific models. The latter uses as input the oracle segmentation of the résumé. Results are in micro-average $F_{1}$ scores for the positive labels, aggregated by entity type of each section: \textit{contact information} (\textbf{Cont}), \textit{work experience} (\textbf{Work}), and \textit{education} (\textbf{Edu}). In each case, the average result of three independent replications using different random seeds is reported.}
    \label{tab:results-goals-separate}}
    \centering
    \begin{tabular}{lcccccccccccccc}
    \hline
    \Tsxx \multirow{2}{*}{\textbf{Model}} & & \multicolumn{3}{c}{\textbf{English}} & & \multicolumn{3}{c}{\textbf{French}} & & \multicolumn{3}{c}{\textbf{Chinese}} \\ \cline{3-5} \cline{7-9} \cline{11-13} 
                                 & & Cont. & Work & Edu. & & Cont. & Work & Edu. & & Cont. & Work & Edu. \\ 
    \hline
    \Tsxx \textit{FastText features} & & & & & & & & & & & & \\
    ~~Multi-task (lines and tokens) & & 
    93.43 &   
    88.12 &   
    83.35 & & 
    96.29 &   
    83.96 &   
    85.95 & & 
    97.35 &   
    89.93 &   
    93.33     
        \\
    ~~Section-specific (only tokens) & &
    95.79 &   
    90.62 &   
    88.33 & & 
    95.90 &   
    86.30 &   
    88.77 & & 
    98.31 &   
    92.21 &   
    94.79     
        \\
   \T \textit{Transformer features} & & & & & & & & & & & & \\
    ~~Multi-task (lines and tokens) & &
    95.44 &   
    90.32 &   
    86.30 & & 
    95.83 &   
    86.98 &   
    90.06 & & 
    95.92 &   
    89.36 &   
    94.15     
        \\
    ~~Section-specific (only tokens) & &
    94.19 &   
    91.55 &   
    88.49 & & 
    94.77 &   
    86.22 &   
    90.50 & & 
    96.15 &   
    92.64 &   
    95.33     
        \\
    \hline
    \end{tabular}
\end{table*}


\subsubsection*{Ablations and Comparison with Previous Work}

Table~\ref{tab:results-ablation} presents an ablation study of the proposed architectures in order to empirically support our architectural design choices. Furthermore, some of the ablated variants are re-implementations of systems proposed in previous work and thus act as baselines for the experiments presented above in this section.

The first group involves variants that use, as initial features, the combination of FastText word embeddings and handcrafted features. Variant~1 is the multi-task model presented in Table~\ref{tab:results-goals-main}. The first ablation, variant~2, involves replacing the top-wise CRF layer with a Softmax layer. Both variants have comparable performance, with a small degradation when Softmax is used. 
The next ablation, variant~3, removes the BiRNN layer and thus makes the CRF predict the token labels using the initial features directly. This is a re-implementation of the system proposed by Tosik et al.~(2015)~\cite{tosik-etal-2015-word}, although they did not share their handcrafted features (and therefore we use those described in Appendix~\ref{sec:handcrafted}). This other ablated variant has a substantial degradation in performance with respect to our proposed model, suggesting that the role played by the BiRNN layer is critical.

The second group involves variants that apply frozen Transformers to each line individually, and then concatenate every line to obtain the initial features (this is visually described in Figure~\ref{fig:architecture-detail}). Variant~4 is the multi-task model presented in Table~\ref{tab:results-goals-main}. The first ablation, variant~5, involves replacing the T5 (or mT5) encoder with a BERT (or mBERT) encoder~\cite{devlin-etal-2019-bert}. We observe an appreciable degradation in performance, suggesting that the pre-trained T5 family of models produces representations that are more useful for our task. Variant~6~and~7 use T5 and BERT, respectively, but omit the recurrent layer. Both result in a significant degradation of performance with respect to the models including the BiRNN, again showing the importance of the BiRNN for this task.

The third group involves variants that also apply Transformers to each individual line, but this time we allow for the Transformer encoder to be fine-tuned with the task supervision. In this case, we do not employ a BiRNN for contextualizing token representations across lines because this would require a much more challenging optimization procedure\footnote{A naïve implementation for this procedure would require keeping in memory as many copies of the Transformer as lines in the target résumé.} and thus each line is processed independently. Variant~8 involves a BERT encoder (being fine-tuned) that computes representations for each token in the line, and uses a CRF layer to predict their labels. When compared to our proposed model (variant~4), we observe a significant drop in performance, suggesting that the contextualization across different lines in the résumé is the critical factor for the performance of the system. Interestingly, when variant~8 is compared to variant~7 ---identical, except for fine-tuning--- we do see an improvement in performance, suggesting that without inter-line contextualization, fine-tuning is indeed helpful.

Variant~9 is similar to the previous variant but replaces the CRF layer with Softmax. This model is a re-implementation of the NER system presented by Devlin~et~al.~(2019)~\cite{devlin-etal-2019-bert} and it is also equivalent to the system used for résumé parsing by Barducci~et~al.~(2022)~\cite{barducci2022ir}. We can observe that the performance is similar to the previous one (although CRF seems to achieve slightly better results). Lastly, variant~10 is included as a control, in which we fine-tune the encoder component of T5 and predict labels using a Softmax on top of the Transformer representations for each token. Again, T5 provides slightly better results with respect to the equivalent BERT variant.

In summary, the ablation experiments suggest that the BiRNN layer, which contextualizes the token representations across the entire résumé, has a significant impact on the performance. The CRF helps to further improve the performance but in a smaller amount. The variants that allow for fine-tuning the Transformer component outperform their frozen-Transformer equivalents, but they are in turn outperformed by our proposed solutions (variants~1~and~4).

\begin{table}[t]
    \setlength{\tabcolsep}{5pt}
    \caption{Ablation study. Variants are compared in terms of the micro-average $F_{1}$ obtained for the token sequence labeling task. Variants~1~and~4 represent the models discussed in the previous part of this section. Other model variants depart from either one of these by changing one aspect at a time. In particular, variant~3 re-implements the system of Tosik~et~al.~(2015)~\cite{tosik-etal-2015-word}, and variant~9 is equivalent to the architecture proposed by Devlin~et~al.~(2019)~\cite{devlin-etal-2019-bert} for other sequence labeling tasks. \textit{IF} denotes \textit{initial features}. Each result is an average of three independent replications.}
    \label{tab:results-ablation}
    \centering
    \begin{tabular}{lccc}
    \hline
    \textbf{Model variant} & \textbf{English} & \textbf{French} & \textbf{Chinese} \\
    \hline
    
    \multicolumn{4}{l}{\Tsxx \textit{FastText initial features}} \\ 
    1 IF+BiRNN+CRF           & 89.03   & 86.90  & 92.66   \\
    2 IF+BiRNN+Softmax       & 88.86   & 86.53  & 92.67   \\
    3 IF+CRF \cite{tosik-etal-2015-word} & 65.89   & 64.68  & 67.53   \\
    
    \multicolumn{4}{l}{\T  \textit{Transformer initial features}} \\
    \multicolumn{4}{l}{\textit{(frozen)}} \\
    4 T5+BiRNN+CRF           & 90.94   & 88.65  & 92.61   \\
    5 BERT+BiRNN+CRF         & 88.91   & 86.34  & 91.79       \\
    6 T5+CRF                 & 78.65   & 75.40  & 76.91   \\
    7 BERT+CRF               & 74.70   & 73.53  & 81.51   \\

    \multicolumn{4}{l}{\T \textit{Transformer initial features, linewise}} \\
    \multicolumn{4}{l}{\textit{(fine-tuned)}} \\
    8 BERT+CRF               & 83.55   & 85.60  & 86.36   \\
    9 BERT+Softmax \cite{devlin-etal-2019-bert, barducci2022ir} & 83.13   & 85.55  & 85.95   \\
    10 T5+Softmax             & 84.18   & 85.61      & 86.58   \\

    \hline
\end{tabular}
\end{table}


\section{Conclusion}
\label{sec:conclusions}

Résumé parsing is an important task for digitalized recruitment processes, and the accuracy of the parsing step affects downstream recommender systems significantly.

In this work, we study résumé parsing extensively in seven languages. We formulated it as a sequence labeling problem in two levels (lines and tokens), and studied several variants of a unified model that solves both tasks. We also described the process for developing high-quality annotated corpora in seven languages. We showed through experimental results that the proposed models can perform this task effectively despite the challenges of substantially long input text sequences and a large number of labels. We observed that the joint model is more convenient than the typical two-stage solution in terms of resource efficiency and model life-cycle maintainability, and also found that in some cases the joint model yields better performance. We provided a trade-off analysis of the proposed variants and described challenges for deployment in production environments. The ablation experiments suggest that the BiRNN layer contextualizing across the résumé is critical for performance, and that the CRF component further provides a smaller improvement.

Potential directions for future research include the following: using character-based initial features~\cite{lample-etal-2016-neural, akbik-etal-2018-contextual} for the FastText variants, as they can complement word embeddings by incorporating information from the surface form of the text and may even offer the opportunity to gradually replace handcrafted features; domain-adapting the Transformer representations with unannotated résumés, considering the reported effectiveness of this technique in enhancing downstream task performance~\cite{gururangan-etal-2020-dont}; and building multilingual models to improve sample efficiency for low-resource languages. Furthermore, alternative Transformer architectures designed specifically for long input sequences~\cite{dai-etal-2019-transformer, hutchins2022} could be used in order to encode the entire résumé in a single pass, while also enabling the possibility to fine-tune the encoder.


\section*{Limitations}
\label{sec:limitations}

As discussed in Section~\ref{sec:corpora}, despite our best efforts to cover as many locations, industries, and seniority levels, it is not feasible for résumé parsing corpora with sizes of up to 1200 résumés to actually contain samples from every subgroup of the population under study. Therefore, we would like to highlight that the findings presented in this work apply specifically to résumés that are similar to those included in the corpora, and may not generalize with the same level of accuracy to other résumés belonging to combinations of location, industry, and work experience that were not seen by the model during training.


\section*{Ethics Statement}
\label{sec:ethics}

The system described in this work is intended for parsing résumés of individuals from different backgrounds, located around the globe. Considering the importance of inclusivity in this context, we made a great effort to cover the diversity of the use of language in our corpora with the objective in mind. This helps us to provide high-quality résumé parsing for individuals from various industries and locations.

Furthermore, the data used for training and evaluating our models consist of résumés that contain sensitive information from real-world individuals. We have taken the necessary privacy and security measures for protecting this information throughout every step of this project.

\bibliography{referenes}

\appendix


\section{Details on the Corpus Annotations}
\label{sec:appendix-guidelines}

\begin{figure*}[t]
    \centering
    \includegraphics[width=0.75\textwidth]{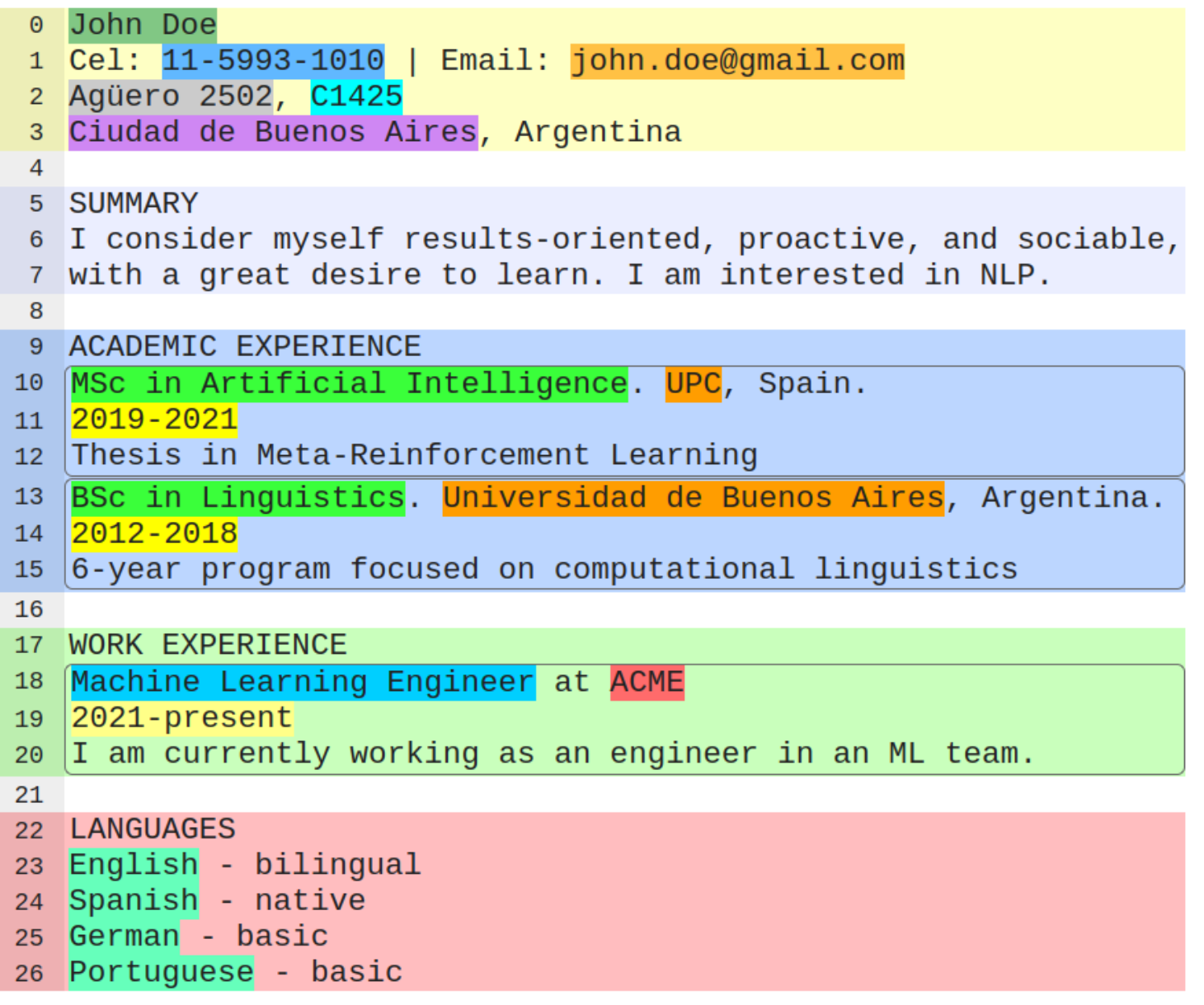}
    \caption{Screenshot of the user interface of the custom annotation tool, showing the plain-text version of a fictional résumé with its corresponding annotations.}
    \centering
    \label{fig:annotations-tool}
\end{figure*}

Annotators were provided access to a custom annotation tool that we developed for this task. In Figure~\ref{fig:annotations-tool} we show the user interface of this tool, with an example résumé annotated for sections, groups, and entities.

The annotators were asked to detect and highlight specific information in résumés written in their language. They were introduced to the different types of information they would be annotating: line annotations include sections and groups, and text annotations (allowing for any span of characters in the text) include entities.

The description for every type of annotation label is included below. 

\subsection{Labels for Lines}

\label{subsec:appendix-guidelines-lines}

The sequence labeling task at the line-level is intended to recognize sections and groups within certain sections. The label set includes a total of 18 different labels.


\subsubsection{Sections}

We allow the annotators to label lines into the following sections:

\begin{quote}
{\scriptsize
\begin{description}
\item[Contact Information] Contact and personal information about the candidate.
\item[Work Experience] Information about the candidate's employment experience.
\item[Education] Information about the candidate's formal education. 
\item[Internship] Information about the candidate's internship experience.
\item[Skills] Information about the candidate's work-related abilities and qualifications.
\item[Languages] A description of the language proficiency of the candidate.
\item[Summary] Brief statement intended to display a candidate’s most compelling abilities and attributes.
\item[Objective] Candidate's work-related goals, usually written in prose. It emphasizes what the person is looking for from a job or company.
\item[Achievements] It includes content of importance under titles like Honors, Awards, Accomplishments, Publications, Licenses and Certifications, etc.
\item[References] Contact details of individuals who can speak about the character, academic work or experience, and extracurricular achievements of the candidate
\item[Letter] Letters embedded in the résumé (frequently cover letters and reference letters).
\end{description}
}
\end{quote}

For sections, we use IO (inside, outside) format for the labels. This is motivated by the fact that we assume that two consecutive lines of the same type of section belong to the same section element.


\subsubsection{Groups}

Besides the categories listed above, lines that belong to sections \textbf{Work Experience}, \textbf{Education}, and \textbf{Internship} can belong to experience groups. Internally, the tool uses the IOB (inside, outside, beginning) format for the groups within each section type. Note that, for example in the case of the Education section, the label \texttt{I-edu} denotes a line that is part of the Education section but it's not part of any particular group, whereas the label \texttt{B-edu\_group} denotes a line that lies at the beginning of a group in the Education section.


\subsection{Labels for Tokens}

\label{subsec:appendix-guidelines-tokens}

The sequence labeling task at the token-level is intended to extract entities. Entities are annotated on arbitrary spans of characters in the résumé text, in order to generalize the annotations for any possible tokenization. We allow for a total of 17 labels. Although most entities are usually found in specific sections, we allow for annotating any entity in any part of the résumé.


\subsubsection{Contact Information entities}

The following entities are usually found in the \textit{contact information} section.

\begin{quote}
{\scriptsize
\begin{description}
    \item[Name] Candidate's name.
    \item[Phone number] Candidate's phone number.
    \item[Email] Candidate's email address.
    \item[St. address] Unit information (apartment, floor) and neighborhood.
    \item[ZipCode] An alphanumeric postal code.
    \item[City] The city the candidate lives in.
    \item[State] First-level geopolitical subdivision where the candidate is located.
\end{description}
}
\end{quote}


\subsubsection{Work Experience and Internship entities}

These entities are usually found in the groups of either the \textit{work experience} or the \textit{internship} sections.

\begin{quote}
{\scriptsize
\begin{description}
    \item[Company] Name of a candidate's employing organization.
    \item[Job title] Title that the employer gave to the candidate while working for the company.
    \item[Period] Period in which the candidate held the position.
\end{description}
}
\end{quote}


\subsubsection{Education entities}

These entities are usually found in the groups of the \textit{education} section.

\begin{quote}
{\scriptsize
\begin{description}
    \item[School name] Name of an institution where the candidate was formally educated. 
    \item[Degree title] he name of the academic program in which a student participates or was awarded a degree.
    \item[Degree Period] Period of time in which the candidate attended the educational organization in fulfillment of the degree.
    \item[Major] The core academic discipline the candidate had to focus on while pursuing their degree.
    \item[GPA] Grade, or any other candidate score, presented as a standardized measurement.
\end{description}
}
\end{quote}


\subsubsection{Language entities}

The following entities are usually found in the Languages section.

\begin{quote}
{\scriptsize
\begin{description}
    \item[Language name] The name of a language that the candidate claims to be familiar with.
\end{description}
}
\end{quote}


\section{Detail on the Handcrafted Features}
\label{sec:handcrafted}

The non-Transformer models presented in Section~\ref{sec:model} employ a combination of FastText word embeddings and handcrafted features. The complete list and details of each handcrafted feature is provided in GitHub because of space considerations\footnote{\url{https://github.com/federetyk/resume-parsing}}. The final set of features was determined based on the empirical results of preliminary experiments, which are not included in this study due to space constraints. Note that certain features necessitate dictionaries of relevant terms to be computed, thereby requiring separate dictionaries for each language.

\end{document}